\documentclass[preprint,12pt,authoryear]{elsarticle}

\usepackage{acronym}
\usepackage{amsmath}
\usepackage{graphicx}
\usepackage[utf8]{inputenc}
\usepackage{mathtools}
\usepackage{multirow}
\usepackage{tabularx}

\journal{arXiv}

\begin{document}

\newcolumntype{L}[1]{>{\raggedright\arraybackslash}p{#1}}
\newcolumntype{C}[1]{>{\centering\arraybackslash}p{#1}}
\newcolumntype{R}[1]{>{\raggedleft\arraybackslash}p{#1}}

\newacro{AP}{Average Precision}
\newacro{ATL}{Anterior Temporal Lobe}
\newacro{CNN}{Convolutional Neural Network}
\newacro{CCA}{Canonical Correlation Analysis}
\newacro{CSC}{Controlled Semantic Cognition}
\newacro{DCCA}{Deep Canonical Correlation Analysis}
\newacro{EMC}{Embodied Music Cognition}
\newacro{FFT}{Fast Fourier Transform}
\newacro{fMRI}{functional Magnetic Resonance Imaging}
\newacro{GAN}{Generative Adversarial Network}
\newacro{MAP}{Mean Average Precision}
\newacro{MVNN}{Multi-view Neural Network}
\newacro{NTTL}{Neural Theory of Thought and Language}
\newacro{PC}{Predictive Coding}

\begin{frontmatter}

\title{Learning embodied semantics via music and dance semiotic correlations}



\author[1,3]{Francisco Afonso Raposo}
\cortext[cor1]{Corresponding author:
  Tel.: +351-213-100-313}
\ead{francisco.afonso.raposo@tecnico.ulisboa.pt}
\author[1,3]{David Martins de Matos}
\author[2,3]{Ricardo Ribeiro}

\address[1]{Instituto Superior Técnico, Universidade de Lisboa, Av. Rovisco Pais, 1049-001 Lisboa, Portugal}
\address[2]{Instituto Universitário de Lisboa (ISCTE-IUL), Av. das Forças Armadas, 1649-026 Lisboa, Portugal}
\address[3]{INESC-ID Lisboa, R. Alves Redol 9, 1000-029 Lisboa, Portugal}


\begin{abstract}
Music semantics is embodied, in the sense that meaning is biologically mediated by and grounded in the human body and brain. This embodied cognition perspective also explains why music structures modulate kinetic and somatosensory perception. We leverage this aspect of cognition, by considering dance as a proxy for music perception, in a statistical computational model that learns semiotic correlations between music audio and dance video. We evaluate the ability of this model to effectively capture underlying semantics in a cross-modal retrieval task. Quantitative results, validated with statistical significance testing, strengthen the body of evidence for embodied cognition in music and show the model can recommend music audio for dance video queries and vice-versa.
\end{abstract}


\end{frontmatter}

\section{Introduction}

Recent developments in human embodied cognition posit a learning and understanding mechanism called ``conceptual metaphor'' \citep{lakoff2012}, where knowledge is derived from repeated patterns of experience. Neural circuits in the brain are substrates for these metaphors \citep{lakoff2014} and, therefore, are the drivers of semantics. Semantic grounding can be understood as the inferences which are instantiated as activation of these learned neural circuits. While not using the same abstraction of conceptual metaphor, other theories of embodied cognition also cast semantic memory and inference as encoding and activation of neural circuitry, differing only in terms of which brain areas are the core components of the biological semantic system \citep{kiefer2012,ralph2017}. The common factor between these accounts of embodied cognition is the existence of transmodal knowledge representations, in the sense that circuits are learned in a modality-agnostic way. This means that correlations between sensory, motor, linguistic, and affective embodied experiences create circuits connecting different modality-specific neuron populations. In other words, the statistical structure of human multimodal experience, which is captured and encoded by the brain, is what defines semantics. Music semantics is no exception, also being embodied and, thus, musical concepts convey meaning in terms of somatosensory and motor concepts \citep{koelsch2019,kreyn2018,leman2014}.

The statistical and multimodal imperative for human cognition has also been hinted at, at least in some form, by research across various disciplines, such as in aesthetics \citep{cook2000,davies1994,kivy1980,kurth1991,scruton1997}, semiotics \citep{azcarate2011,bennett2008,blanariu2013,lemke1992}, psychology \citep{brown2011,dehaene2007,eitan2006,eitan2011,frego1999,krumhansl1997,larson2004,roffler1968,sievers2013,silver2007,styns2007,wagner1981}, and neuroscience \citep{fujioka2012,janata2012,koelsch2019,nakamura1999,nozaradan2011,penhune1998,platel1997,spence1997,stein1995,widmann2004,zatorre1994}, namely, for natural language, music, and dance. In this work, we are interested in the semantic link between music and dance (movement-based expression). Therefore, we leverage this multimodal aspect of cognition by modeling expected semiotic correlations between these modalities. These correlations are expected because they are mainly surface realizations of cognitive processes following embodied cognition. This framework implies that there is a degree of determinism underlying the relationship between music and dance, that is, dance design and performance are heavily shaped by music. This evident and intuitive relationship is even captured in some natural languages, where words for music and dance are either synonyms or the same \citep{baily1985}. In this work, we claim that, just like human semantic cognition is based on multimodal statistical structures, joint semiotic modeling of music and dance, through statistical computational approaches, is expected to provide some light regarding the semantics of these modalities as well as provide intelligent technological applications in areas such as multimedia production. That is, we can automatically learn the symbols/patterns (semiotics), encoded in the data representing human expression, which correlate across several modalities. Since this correlation defines and is a manifestation of underlying cognitive processes, capturing it effectively uncovers semantic structures for both modalities.

Following the calls for technological applications based on sensorimotor aspects of semantics \citep{leman2010,matyja2016}, this work leverages semiotic correlations between music and dance, represented as audio and video, respectively, in order to learn latent cross-modal representations which capture underlying semantics connecting these two modes of communication. These representations are quantitatively evaluated in a cross-modal retrieval task. In particular, we perform experiments on a 592 music audio-dance video pairs dataset, using \acp{MVNN}, and report 75\% rank accuracy and 57\% pair accuracy instance-level retrieval performances and 26\% \ac{MAP} class-level retrieval performance, which are all statistically very significant effects (p-values $<0.01$). We interpret these results as further evidence for embodied cognition-based music semantics. Potential end-user applications include, but are not limited to, the automatic retrieval of a song for a particular dance or choreography video and vice-versa. To the best of our knowledge, this is the first instance of such a joint music-dance computational model, capable of capturing semantics underlying these modalities and providing a connection between machine learning of these multimodal correlations and embodied cognition perspectives.

The rest of this paper is structured as follows: Section \ref{sec:related} reviews related work on embodied cognition, semantics, and semiotics, motivating this approach based on evidence taken from research in several disciplines; Section \ref{sec:setup} details the experimental setup, including descriptions of the evaluation task, \ac{MVNN} model, dataset, features, and preprocessing; Section \ref{sec:results} presents the results; Section \ref{sec:discussion} discusses the impact of these results; and Section \ref{sec:conclusions} draws conclusions and suggests future work.

\section{Related work}
\label{sec:related}

Conceptual metaphor \citep{lakoff2012} is an abstraction used to explain the relational aspect of human cognition as well as its biological implementation in the brain. Experience is encoded neurally and frequent patterns or correlations encountered across many experiences define conceptual metaphors. That is, a conceptual metaphor is a link established in cognition (often subconsciously) connecting concepts. An instance of such a metaphor implies a shared meaning of the concepts involved. Which metaphors get instantiated depends on the experiences had during a lifetime as well as on genetically inherited biological primitives (which are also learned based on experience, albeit across evolutionary time scales). These metaphors are physically implemented as neural circuits in the brain which are, therefore, also learned based on everyday experience. The learning process at the neuronal level of abstraction is called ``Hebbian learning'', where ``neurons that fire together, wire together'' is the motto \citep{lakoff2014}. Semantic grounding in this theory, called \ac{NTTL}, which is understood as the set of semantic inferences, manifests in the brain as firing patterns of the circuits encoding such metaphorical inferences. These semantics are, therefore, transmodal: patterns of multimodal experience dictate which circuits are learned. Consequently, semantic grounding triggers multimodal inferences in a natural, often subconscious, way. Central to this theory is the fact that grounding is rooted in primitive concepts, that is, inference triggers the firing of neuron populations responsible for perception and action/coordination of the material body interacting in the material world. These neurons encode concepts like movement, physical forces, and other bodily sensations, which are mainly located in the somatosensory and sensorimotor systems \citep{desai2011,guevara2018,koelsch2019,lakoff2014}. Other theories, such as the \ac{CSC} \citep{ralph2017}, share this core multimodal aspect of cognition but defend that a transmodal hub is located in the \acp{ATL} instead. \cite{kiefer2012} review and compare several semantic cognition theories and argue in favor of the embodiment views of conceptual representations, which are rooted in transmodal integration of modality-specific (e.g., sensory and motor) features. In the remainder of this section, we review related work providing evidence for the multimodal nature of cognition and the primacy of primitive embodied concepts in music.

Aesthetics suggests that musical structures evoke emotion through isomorphism with human motion \citep{cook2000,davies1994,kivy1980,scruton1997} and that music is a manifestation of a primordial ``kinetic energy'' and a play of ``psychological tensions'' \citep{kurth1991}. \cite{blanariu2013} claims that, even though the design of choreographies is influenced by culture, its aesthetics are driven by ``pre-reflective'' experience, i.e., unconscious processes driving body movement expression. The choreographer interprets the world (e.g., a song), via ``kinetic thinking'' \citep{laban1960}, which is materialized in dance in such a way that its surface-level features retain this ``motivating character'' or ``invoked potential'' \citep{peirce1991}, i.e., the conceptual metaphors behind the encoded symbols can still be accessible. The symbols range from highly abstract cultural encodings to more concrete patterns, such as movement patterns in space and time such as those in abstract (e.g., non-choreographed) dance \citep{blanariu2013}. \cite{bennett2008} characterizes movement and dance semantics as being influenced by both physiological, psychological, and social factors and based on space and forces primitives. In music, semantics is encoded symbolically in different dimensions (such as timbral, tonal, and rhythmic) and levels of abstraction \citep{juslin2013,schlenker2017}. These accounts of encoding of meaning imply a conceptual semantic system which supports several denotations \citep{blanariu2013}, i.e., what was also termed an ``underspecified'' semantics \citep{schlenker2017}. The number of possible denotations for a particular song can be reduced when considering accompanying communication channels, such as dance, video, and lyrics \citep{schlenker2017}. Natural language semantics is also underspecified according to this definition, albeit to a much lower degree. Furthermore, \cite{azcarate2011} emphasizes the concept of ``intertextuality'' as well as text being a ``mediator in the semiotic construction of reality''. Intertextuality refers to the context in which a text is interpreted, allowing meaning to be assigned to text \citep{lemke1992}. This context includes other supporting texts but also history and culture as conveyed by the whole range of semiotic possibilities, i.e., via other modalities \citep{lemke1992}. That is, textual meaning is also derived via multimodal inferences, which improve the efficacy of communication. This ``intermediality'' is a consequence of human cognitive processes based on relational thinking (conceptual metaphor) that exhibit a multimodal and contextualized inferential nature \citep{azcarate2011}. \cite{peirce1991} termed this capacity to both encode and decode symbols, via semantic inferences, as ``abstractive observation'', which he considered to be a feature required to learn and interpret by means of experience, i.e., required for being an ``intelligent consciousness''.

Human behaviour reflects this fundamental and multimodal aspect of cognition, as shown by psychology research. For instance, \cite{eitan2011} found several correlations between music dimensions and somatosensory-related concepts, such as sharpness, weight, smoothness, moisture, and temperature. People synchronize walking tempo to the music they listen to and this is thought to indicate that the perception of musical pulse is internalized in the locomotion system \citep{styns2007}. The biological nature of the link between music and movement is also suggested in studies that observed pitch height associations with vertical directionality in 1-year old infants \citep{wagner1981} and with perceived spatial elevation in congenitally blind subjects and 4- to 5-year old children who did not verbally make those associations \citep{roffler1968}. Tension ratings performed by subjects independently for either music or a corresponding choreography yielded correlated results, suggesting tension fluctuations are isomorphically manifested in both modalities \citep{frego1999,krumhansl1997}. \cite{silver2007} showed that the perception of ``beat'' is transferable across music and movement for humans as young as 7 months old. \cite{eitan2006} observed a kind of music-kinetic determinism in an experiment where music features were consistently mapped onto kinetic features of visualized human motion. \cite{sievers2013} found further empirical evidence for a shared dynamic structure between music and movement in a study that leveraged a common feature between these modalities: the capacity to convey affective content. Experimenters had human subjects independently control the shared parameters of a probabilistic model, for generating either piano melodies or bouncing ball animations, according to specified target emotions: angry, happy, peaceful, sad, and scared. Similar emotions were correlated with similar slider configurations across both modalities and different cultures: American and Kreung (in a rural Cambodian village which maintained a high degree of cultural isolation). The authors argue that the isomorphic relationship between these modalities may play an important role in evolutionary fitness and suggest that music processing in the brain ``recycles'' \citep{dehaene2007} other areas evolved for older tasks, such as spatiotemporal perception and action \citep{sievers2013}. \cite{brown2011} suggest that this capacity to convey affective content is the reason why music and movement are more cross-culturally intelligible than language. A computational model for melodic expectation, which generated melody completions based on tonal movement driven by physical forces (gravity, inertia, and magnetism), outperformed every human subject, based on intersubject agreement \citep{larson2004}, further suggesting semantic inferences between concepts related to music and movement/forces.

There is also neurological evidence for multimodal cognition and, in particular, for an underlying link between music and movement. Certain brain areas, such as the superior colliculus, are thought to integrate visual, auditory, and somatosensory information \citep{spence1997,stein1995}. \cite{widmann2004} observed evoked potentials when an auditory stimulus was presented to subjects together with a visual stimulus that infringed expected spatial inferences based on pitch. The engagement of visuospatial areas of the brain during music-related tasks has also been extensively reported \citep{nakamura1999,penhune1998,platel1997,zatorre1994}. Furthermore, neural entrainment to beat has been observed as $\beta$ oscillations across auditory and motor cortices \citep{fujioka2012,nozaradan2011}. Moreover, \cite{janata2012} found a link between the feeling of ``being in the groove'' and sensorimotor activity. \cite{kreyn2018} also explains music semantics from an embodied cognition perspective, where tonal and temporal relationships in music artifacts convey embodied meaning, mainly via modulation of physical tension. These tonal relationships consist of manipulations of tonal tension, a core concept in musicology, in a tonal framework (musical scale). Tonal tension is physically perceived by humans as young as one-day-old babies \citep{virtala2013}, which further points to the embodiment of music semantics, since tonal perception is mainly biologically driven. The reason for this may be the ``principle of least effort'', where consonant sounds consisting of more harmonic overtones are more easily processed and compressed by the brain than dissonant sounds, creating a more pleasant experience \citep{bidelman2009,bidelman2011}. \cite{leman2007} also emphasizes the role of kinetic meaning as a translator between structural features of music and semantic labels/expressive intentions, i.e., corporeal articulations are necessary for interpreting music. Semantics are defined by the mediation process when listening to music, i.e., the human body and brain are responsible for mapping from the physical modality (audio) to the experienced modality \citep{leman2010}. This mediation process is based on motor patterns which regulate mental representations related to music perception. This theory, termed \ac{EMC}, also supports the idea that semantics is motivated by affordances (action), i.e., music is interpreted in a (kinetic) way that is relevant for functioning in a physical environment. Furthermore, \ac{EMC} also states that decoding music expressiveness in performance is a sense-giving activity \citep{leman2014}, which falls in line with the learning nature of \ac{NTTL}. The \ac{PC} framework of \cite{koelsch2019} also points to the involvement of transmodal neural circuits in both prediction and prediction error resolution (active inference) of musical content. The groove aspect of music perception entails an active engagement in terms of proprioception and interoception, where sensorimotor predictions are inferenced (by ``mental action''), even without actually moving. In this framework, both sensorimotor and autonomic systems can also be involved in resolution of prediction errors.

Recently, \cite{pereira2018} proposed a method for decoding neural representations into statistically-modeled semantic dimensions of text. This is relevant because it shows statistical computational modeling (in this instance, ridge regression) is able to robustly capture language semantics in the brain, based on \ac{fMRI}. This language-brainwaves relationship is an analogue to the music-dance relationship in this work. The main advantage is that, theoretically, brain activity will directly correlate to stimuli, assuming we can perfectly decode it. Dance, however, can be viewed as an indirect representation, a kinetic proxy for the embodied meaning of the music stimulus, which is assumed to be encoded in the brain. This approach provides further insights motivating embodied cognition perspectives, in particular, to its transmodal aspect. \ac{fMRI} data was recorded for three different text concept presentation paradigms: using it in a sentence, pairing it with a descriptive picture, and pairing it with a word cloud (several related words). The best decoding performance across individual paradigms was obtained with the data recorded in the picture paradigm, illustrating the role of intermediality in natural language semantics and cognition in general. Moreover, an investigation into what voxels were most informative for decoding, revealed that they were from widely distributed brain areas (language 21\%, default mode 15\%, task-positive 23\%, visual 19\%, and others 22\%), as opposed to being focalized in the language network, further suggesting an integrated semantic system distributed across the whole brain. A limitation of that approach in relation to the one proposed here is that regression is performed for each dimension of the text representation independently, failing to capture how all dimensions jointly covary across both modalities.

\section{Experimental setup}
\label{sec:setup}

As previously stated, multimedia expressions referencing the same object (e.g., audio and dance of a song) tend to display semiotic correlations reflecting embodied cognitive processes. Therefore, we design an experiment to evaluate how correlated these artifact pairs are: we measure the performance of cross-modal retrieval between music audio and dance video. The task consists of retrieving a sorted list of relevant results from one modality, given a query from another modality. We perform experiments in a 4-fold cross-validation setup and report pair and rank accuracy scores (as done by \cite{pereira2018}) for instance-level evaluation and \ac{MAP} scores for class-level evaluation. The following sections describe the dataset (Section \ref{sub:dataset}), features (Section \ref{sub:features}), preprocessing (Section \ref{sub:preprocessing}), \ac{MVNN} model architecture and loss function (Section \ref{sub:model}), and evaluation details (Section \ref{sub:evaluation}).

\subsection{Dataset}
\label{sub:dataset}

We ran experiments on a subset of the \emph{Let's Dance} dataset of 1000 videos of dances from 10 categories: ballet, breakdance, flamenco, foxtrot, latin, quickstep, square, swing, tango, and waltz \citep{castro2018}. This dataset was created in the context of dance style classification based on video. Each video is 10s long and has a rate of 30 frames per second. The videos were taken from YouTube at 720p quality and include both dancing performances and practicing. We used only the audio and pose detection data (body joint positions) from this dataset, which was extracted by applying a pose detector \citep{wei2016} after detecting bounding boxes in a frame with a real-time person detector \citep{redmon2016}. After filtering out all instances which did not have all pose detection data for 10s, the final dataset size is 592 pairs.

\subsection{Features}
\label{sub:features}


The audio features consist of logarithmically scaled Mel-spectrograms extracted from 16,000Hz audio signals. Framing is done by segmenting chunks of 50ms of audio every 25ms. Spectra are computed via \ac{FFT} with a buffer size of 1024 samples. The number of Mel bins is set to 128, which results in a final matrix of 399 frames by 128 Mel-frequency bins per 10s audio recording. We segment each recording into 1s chunks (50\% overlap) to be fed to the \ac{MVNN} (detailed in Section \ref{sub:model}), which means that each of the 592 objects contains 19 segments (each containing 39 frames), yielding a dataset of a total of 11,248 samples.

\begin{figure}
\includegraphics[width=\linewidth]{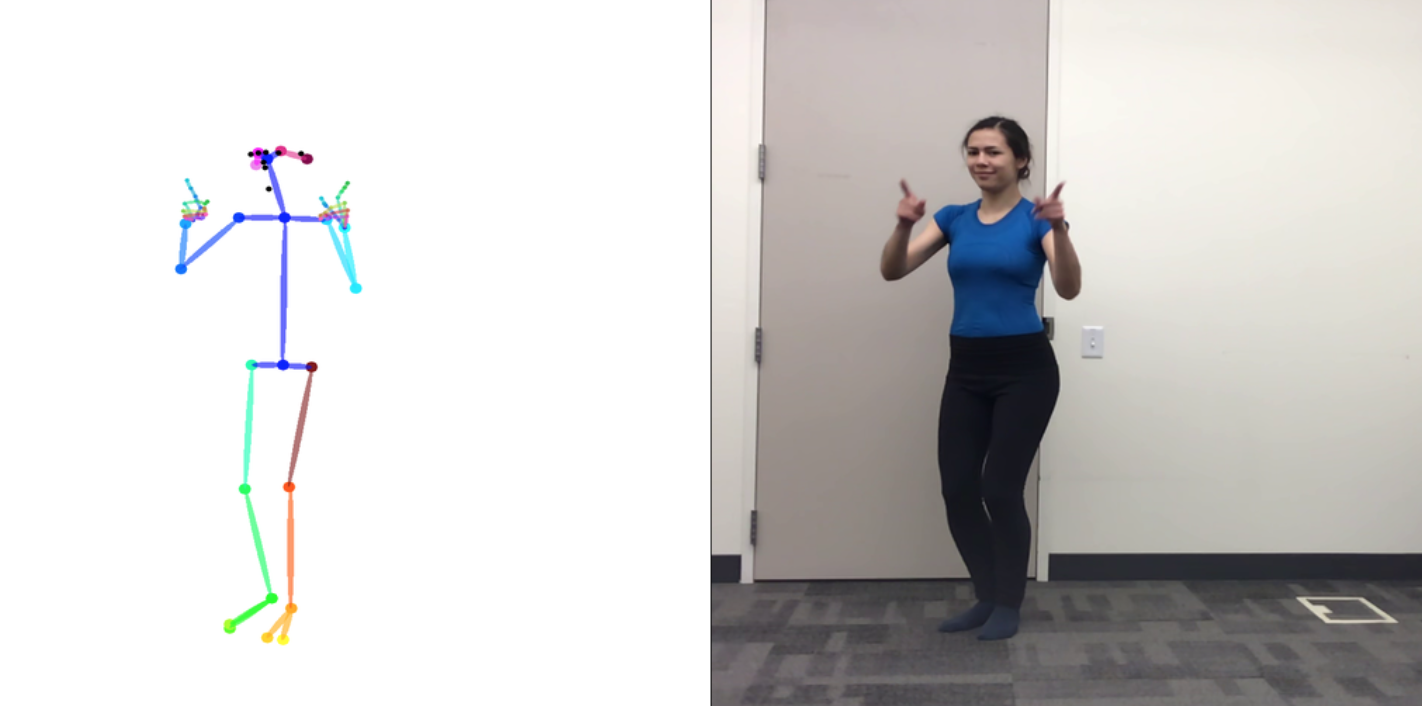}
\caption{Pose detection illustration taken from \citep{chan2018}. Skeleton points represent joints.}
\label{fig:keypoints}
\end{figure}

The pose detection features consist of body joint positions in frame space, i.e., pixel coordinates ranging from 0 (top left corner) to 1280 and 720 for width and height, respectively. The positions for the following key points are extracted: head, neck, shoulder, elbow, wrist, hip, knee, and ankle. There are 2 keypoints, left and right, for each of these except for head and neck, yielding a total of 28 features (14 keypoints with 2 coordinates, $\operatorname{x}$ and $\operatorname{y}$, each). Figure \ref{fig:keypoints} illustrates the keypoints.
These features are extracted at 30fps for the whole 10s video duration ($t\in \{t_0 ... t_{299}\}$), normalized after extraction according to Section \ref{sub:preprocessing}, and then derived features are computed from the normalized data. The position and movement of body joints are used together for expression in dance. Therefore, we compute features that reflect the relative positions of body joints in relation to each other. This translates into computing the euclidean distance between each combination of two joints, yielding 91 derived features and a total of 119 movement features. As for audio, we segment this sequence into 1s segments (50\% overlap), each containing 30 frames.

\subsection{Preprocessing}
\label{sub:preprocessing}

We are interested in modeling movement as bodily expression. Therefore, we should focus on the temporal dynamics of joint positions relative to each other in a way that is as viewpoint- and subject-invariant as possible. However, the positions of subjects in frame space varies according to their distance to the camera. Furthermore, limb proportions are also different across subjects. Therefore, we normalize the joint position data in a similar way to \cite{chan2018}, whose purpose was to transform a pose from a source frame space to a target frame space. We select an arbitrary target frame and project every source frame to this space. We start by taking the maximum ankle $\tt{y}$ coordinate $\tt{ankl}^{\tt{clo}}$ (Equation \ref{eq:clo}) and the maximum ankle $\tt{y}$ coordinate which is smaller than (spatially above) the median ankle $\tt{y}$ coordinate $\tt{ankl}^{\tt{med}}$ (Equation \ref{eq:med}) and about the same distance to it as the distance between it and $\tt{ankl}^{\tt{clo}}$ ($\tt{ankl}^{\tt{far}}$ in Equation \ref{eq:far}). These two keypoints represent the closest and furthest ankle coordinates to the camera, respectively. Formally:
\begin{equation*}
\tt{ankl} = \{\tt{ankl\_y}^{\tt{L}}_t\} \cup \{\tt{ankl\_y}^{\tt{R}}_t\}
\end{equation*}
\begin{equation}
\tt{ankl}^{\tt{clo}} = \operatorname{max}_t(\{\tt{y}_t : \tt{y}_t \in \tt{ankl}\})
\label{eq:clo}
\end{equation}
\begin{equation}
\tt{ankl}^{\tt{med}} = \operatorname{median}_t(\{\tt{y}_t : \tt{y}_t \in \tt{ankl}\})
\label{eq:med}
\end{equation}
\begin{equation}
\tt{ankl}^{\tt{far}} = \operatorname{max}_t(\{\tt{y}_t : \tt{y}_t \in \tt{ankl} \wedge \tt{y}_t < \tt{ankl}^{\tt{med}} \wedge |\tt{y}_t - \tt{ankl}^{\tt{med}}| - \alpha|\tt{ankl}^{\tt{clo}} - \tt{ankl}^{\tt{med}}| < \epsilon\})
\label{eq:far}
\end{equation}
where $\tt{ankl\_y}^{\tt{L}}_t$ and $\tt{ankl\_y}_t^{\tt{R}}$ are the $\tt{y}$ coordinates of the left and right ankles at timestep $\tt{t}$, respectively. Following \citep{chan2018}, we set $\tt{\alpha}$ to 1, and $\tt{\epsilon}$ to 0.7. Then, we computed a scale $s$ (Equation \ref{eq:scale}) to be applied to the $\tt{y}$-axis according to an interpolation between the ratios of the maximum heights between the source and target frames, $\tt{heig}^{\tt{far}}_{src}$ and $\tt{heig}^{\tt{far}}_{tgt}$, respectively. For each dance instance, frame heights are first clustered according to the distance between corresponding ankle $\tt{y}$ coordinate and $\tt{ankl}^{\tt{clo}}$ and $\tt{ankl}^{\tt{far}}$ and then the maximum height values for each cluster are taken (Equations \ref{eq:heigclo} and \ref{eq:heigfar}). Formally:
\begin{equation}
s = \frac{\tt{heig}^{\tt{far}}_{tgt}}{\tt{heig}^{\tt{far}}_{src}} + \frac{\tt{ankl}^{\tt{avg}}_{src} - \tt{ankl}^{\tt{far}}_{src}}{\tt{ankl}^{\tt{clo}}_{src} - \tt{ankl}^{\tt{far}}_{src}} \left(\frac{\tt{heig}^{\tt{clo}}_{tgt}}{\tt{heig}^{\tt{clo}}_{src}} - \frac{\tt{heig}^{\tt{far}}_{tgt}}{\tt{heig}^{\tt{far}}_{src}}\right)
\label{eq:scale}
\end{equation}
\begin{equation}
\tt{heig}^{\tt{clo}} = \operatorname{max}_t(\{|\tt{head\_y}_t - \tt{ankl}^{\tt{LR}}_t| : |\tt{ankl}^{\tt{LR}}_t - \tt{ankl}^{\tt{clo}}| < |\tt{ankl}^{\tt{LR}}_t - \tt{ankl}^{\tt{far}}|\})
\label{eq:heigclo}
\end{equation}
\begin{equation}
\tt{heig}^{\tt{far}} = \operatorname{max}_t(\{|\tt{head\_y}_t - \tt{ankl}^{\tt{LR}}_t| : |\tt{ankl}^{\tt{LR}}_t - \tt{ankl}^{\tt{clo}}| > |\tt{ankl}^{\tt{LR}}_t - \tt{ankl}^{\tt{far}}|\})
\label{eq:heigfar}
\end{equation}
\begin{equation*}
\tt{ankl}^{\tt{LR}}_t = \frac{\tt{ankl\_y}^L_t + \tt{ankl\_y}^R_t}{2}
\end{equation*}
\begin{equation*}
\tt{ankl}^{\tt{avg}} = \operatorname{average}_t(\{\tt{y}_t : \tt{y}_t \in \tt{ankl}\})
\end{equation*}
where $\tt{head\_y}_t$ is the $\tt{y}$ coordinate of the head at timestep $\tt{t}$. After scaling, we also apply a 2D translation so that the position of the ankles of the subject is centered at 0. We do this by subtracting the median coordinates ($\tt{x}$ and $\tt{y}$) of the mean of the (left and right) ankles, i.e., the median of $\tt{ankl}^{\tt{LR}}_t$.

\subsection{Multi-view neural network architecture}
\label{sub:model}

The \ac{MVNN} model used in this work is composed by two branches, each modeling its own view. Even though the final embeddings define a shared and correlated space, according to the loss function, the branches can be arbitrarily different from each other. The loss function is \ac{DCCA} \citep{andrew2013}, a non-linear extension of \ac{CCA} \citep{hotelling1936}, which has also been successfully applied to music by \cite{kelkar2018} and \cite{yu2019}. \ac{CCA} linearly projects two distinct view spaces into a shared correlated space and was suggested to be a general case of parametric tests of statistical significance \citep{knapp1978}. Formally, \ac{DCCA} solves:

\begin{equation}
\left(w_{\bf{x}}^*,w_{\bf{y}}^*,\varphi_{\bf{x}}^*,\varphi_{\bf{y}}^*\right)=\underset{\left(w_{\bf{x}},w_{\bf{y}},\varphi_{\bf{x}},\varphi_{\bf{y}}\right)}{\operatorname{argmax}}\operatorname{corr}\left(w_{\bf{x}}^{\bf{T}}\varphi_{\bf{x}}\left(\bf{x}\right),w_{\bf{y}}^{\bf{T}}\varphi_{\bf{y}}\left(\bf{y}\right)\right)
\end{equation}
where $\bf{x}\in{\rm I\!R}^m$ and $\bf{y}\in{\rm I\!R}^n$ are the zero-mean observations for each view. $\varphi_{\bf{x}}$ and $\varphi_{\bf{y}}$ are non-linear mappings for each view, and $w_{\bf{x}}$ and $w_{\bf{y}}$ are the canonical weights for each view. We use backpropagation and minimize:
\begin{equation}
-\sqrt{\operatorname{tr}\left(\left(C_{XX}^{-1/2}C_{XY}C_{YY}^{-1/2}\right)^{\bf{T}}\left(C_{XX}^{-1/2}C_{XY}C_{YY}^{-1/2}\right)\right)}\
\end{equation}
\begin{equation}
C_{XX}^{-1/2}=Q_{XX}\Lambda_{XX}^{-1/2} Q_{XX}^{\bf{T}}
\end{equation}
where $X$ and $Y$ are the non-linear projections for each view, i.e., $\varphi_{\bf{x}}\left(\bf{x}\right)$ and $\varphi_{\bf{y}}\left(\bf{y}\right)$, respectively. $C_{XX}$ and $C_{YY}$ are the regularized, zero-centered covariances while $C_{XY}$ is the zero-centered cross-covariance. $Q_{XX}$ are the eigenvectors of $C_{XX}$ and $\Lambda_{XX}$ are the eigenvalues of $C_{XX}$. $C_{YY}^{-1/2}$ can be computed analogously. We finish training by computing a forward pass with the training data and fitting a linear \ac{CCA} model on those non-linear mappings. The canonical components of these deep non-linear mappings implement our semantic embeddings space to be evaluated in a cross-modal retrieval task. Functions $\varphi_{\bf{x}}$ and $\varphi_{\bf{y}}$, i.e., the audio and movement projections are implemented as branches of typical neural networks, described in Tables \ref{tab:audio} and \ref{tab:movement}. We use \emph{tanh} activation functions after each convolution layer. Note that other loss functions, such as ones based on pairwise distances \citep{hermann2014,he2017}, can theoretically also be used for the same task. The neural network models were all implemented using TensorFlow \citep{abadi2015}.

\begin{table}[htbp]
\normalsize
\begin{center}
\caption{Audio Neural Network Branch}
\label{tab:audio}
\begin{tabular}{c|C{4mm}cC{4mm}cC{4mm}|c}
\hline
layer type & \multicolumn{5}{c|}{dimensions} & \# params\\
\hline
input & 39 & $\times$ & 128 & $\times$ & 1 & 0\\
\hline
batch norm & 39 & $\times$ & 128 & $\times$ & 1 & 4\\
\hline
2D conv & 39 & $\times$ & 128 & $\times$ & 8 & 200\\
\hline
2D avg pool & 13 & $\times$ & 16 & $\times$ & 8 & 0\\
\hline
batch norm & 13 & $\times$ & 16 & $\times$ & 8 & 32\\
\hline
2D conv & 13 & $\times$ & 16 & $\times$ & 16 & 2064\\
\hline
2D avg pool & 3 & $\times$ & 4 & $\times$ & 16 & 0\\
\hline
batch norm & 3 & $\times$ & 4 & $\times$ & 16 & 64\\
\hline
2D conv & 3 & $\times$ & 4 & $\times$ & 32 & 6176\\
\hline
2D avg pool & 1 & $\times$ & 1 & $\times$ & 32 & 0\\
\hline
batch norm & 1 & $\times$ & 1 & $\times$ & 32 & 128\\
\hline
2D conv & 1 & $\times$ & 1 & $\times$ & 128 & 4224\\
\hline
\multicolumn{6}{c|}{Total params} & 12892\\
\hline
\end{tabular}
\end{center}
\end{table}

\begin{table}[htbp]
\normalsize
\begin{center}
\caption{Movement Neural Network Branch}
\label{tab:movement}
\begin{tabular}{c|C{4mm}cC{4mm}|c}
\hline
layer type & \multicolumn{3}{c|}{dimensions} & \# of params\\
\hline
input & 30 & $\times$ & 119 & 0\\
\hline
batch norm & 30 & $\times$ & 119 & 476\\
\hline
gru & 1 & $\times$ & 32 & 14688\\
\hline
\multicolumn{4}{c|}{Total params} & 15164\\
\hline
\end{tabular}
\end{center}
\end{table}

\subsection{Cross-modal retrieval evaluation}
\label{sub:evaluation}

In this work, cross-modal retrieval consists of retrieving a sorted list of videos given an audio query and vice-versa. We perform cross-modal retrieval on full objects even though the \ac{MVNN} is modeling semiotic correlation between segments. In order to do this, we compute object representations as the average of the \ac{CCA} projections of its segments (for both modalities) and compute the cosine similarity between these cross-modal embeddings. We evaluate the ability of the model to capture semantics and generalize semiotic correlations between both modalities by assessing if relevant cross-modal documents for a query are ranked on top of the retrieved documents list. We define relevant documents in two ways: instance- and class-level. Instance-level evaluation considers the ground truth pairing of cross-modal objects as criterion for relevance, (i.e., the only relevant audio document for a dance video is the one that corresponds to the song that played in that video). Class-level evaluation considers that any cross-modal object sharing some semantic label is relevant (e.g., relevant audio documents for a dance video of a particular dance style are the ones that correspond to songs that played in videos of the same dance style). We perform experiments in a 4-fold cross-validation setup, where each fold partitioning is such that the distribution of classes is similar for each fold. We also run the experiments 10 runs for each fold and report the average performance across runs.

We compute pair and rank accuracies for instance-level evaluation (similar to \cite{pereira2018}). Pair accuracy evaluates ranking performance in the following way: for each query from modality $X$, we consider every possible pairing of the relevant object (corresponding cross-modal pair) and non-relevant objects from modality $Y$. We compute the similarities between the query and each of the two cross-modal objects, as well as the similarities between both cross-modal objects and the corresponding non-relevant object form modality $X$. If the corresponding cross-modal objects are more similar than the alternative, the retrieval trial is successful. We report the average values over queries and non-relevant objects. We also compute a statistical significance test in order to show that the model indeed captures semantics underlying the artifacts. We can think of each trial as a binomial outcome, aggregating two binomial outcomes, where the probability of success for a random model is $0.5\times 0.5=0.25$. Therefore, we can perform a binomial test and compute its p-value. Even though there are $144\times 143$ trials, we consider a more conservative value for the trials $144$ (the number of independent queries). If the p-value is lower than $0.05$, then we can reject the null hypothesis that the results of our model are due to chance. Rank accuracy is the (linearly) normalized rank of the relevant document in the retrieval list: $\operatorname{ra}=1-\left(r-1\right)/\left(L-1\right)$, where $r$ is the rank of the relevant cross-modal object in the list with $L$ elements. This is similar to the pair accuracy evaluation, except that we only consider the query from modality $X$ and the objects from modality $Y$, i.e., each trial consists of one binomial outcome, where the probability of success for a random model is $0.5$. We also consider a conservative binomial test number of trials of $144$ for this metric.

Even though the proposed model and loss function do not explicitly optimize class separation, we expect it to still learn embeddings which capture some aspects of the dance genres in the dataset. This is because different instances of the same class are expected to share semantic structures. Therefore, we perform class-level evaluation, in order to further validate that our model captures semantics underlying both modalities. We compute and report \ac{MAP} scores for each class, separately, and perform a permutation test on these scores against random model performance (whose \ac{MAP} scores are computed according to \cite{bestgen2015}), so that we can show these results are statistically significant and not due to chance. Formally:
\begin{equation}
\operatorname{MAP}_C = \frac{1}{|Q_C|} \sum_{q\in Q_C} \operatorname{AP}_C\left(q\right)
\end{equation}
\begin{equation}
\operatorname{AP}_C\left(q\right) = \frac{\sum_{j=1}^{|R|}\operatorname{pr}\left(j\right)\operatorname{rel}_C\left(r_j\right)}{|R_C|}
\end{equation}
where $C$ is the class, $Q_C$ is the set of queries belonging to class $C$, $\operatorname{AP}_C\left(q\right)$ is the \ac{AP} for query $q$, $R$ is the list of retrieved objects, $R_C$ is the set of retrieved objects belonging to class $C$, $\operatorname{pr}\left(j\right)$ is the precision at cutoff $j$ of the retrieved objects list, and $\operatorname{rel}_C\left(r\right)$ evaluates whether retrieved object $r$ is relevant or not, i.e., whether it belongs to class $C$ or not. Note that the retrieved objects list always contains the whole (train or test) set of data from modality $Y$ and that its size is equal to the total number of (train or test) evaluated queries from modality $X$. \ac{MAP} measures the quality of the sorting of retrieved items lists for a particular definition of relevance (dance style in this work).

\section{Results}
\label{sec:results}

Instance-level evaluation results are reported in Tables \ref{tab:pair} and \ref{tab:rank} for pair and rank accuracies, respectively, for each fold. Values shown in the X / Y format correspond to results when using audio / video queries, respectively. The model was able to achieve 57\% and 75\% for pair and rank accuracies, respectively, which are statistically significantly better (p-values $<0.01$) than the random baseline performances of 25\% and 50\%, respectively.

\begin{table*}[htbp]
\normalsize
\begin{center}
\caption{Instance-level Pair Accuracy}
\label{tab:pair}
\begin{tabular}{cccc|c|c}
\hline
Fold 0 & Fold 1 & Fold 2 & Fold 3 & Average & Baseline\\
\hline
0.57 / 0.57 & 0.57 / 0.56 & 0.60 / 0.59 & 0.55 / 0.56 & 0.57 / 0.57 & 0.25\\
\hline
\end{tabular}
\end{center}
\end{table*}

\begin{table*}[htbp]
\normalsize
\begin{center}
\caption{Instance-level Rank Accuracy}
\label{tab:rank}
\begin{tabular}{cccc|c|c}
\hline
Fold 0 & Fold 1 & Fold 2 & Fold 3 & Average & Baseline\\
\hline
0.75 / 0.75 & 0.75 / 0.75 & 0.77 / 0.76 & 0.74 / 0.74 & 0.75 / 0.75 & 0.50\\
\hline
\end{tabular}
\end{center}
\end{table*}

Class-level evaluation results (\ac{MAP} scores) are reported in Table \ref{tab:map} for each class and fold. The model achieved 26\%, which is statistically significantly better (p-value $<0.01$) than the random baseline performance of 13\%.


\begin{table*}[htbp]
\normalsize
\begin{center}
\caption{Class-level \ac{MAP}}
\label{tab:map}
\begin{tabular}{c|cccc|c|c}
\hline
Style & Fold 0 & Fold 1 & Fold 2 & Fold 3 & Average & Baseline\\
\hline
Ballet & 0.43 / 0.40 & 0.33 / 0.31 & 0.51 / 0.41 & 0.37 / 0.32 & 0.41 / 0.36 & 0.10\\
\hline
Breakdance & 0.18 / 0.17 & 0.18 / 0.14 & 0.18 / 0.14 & 0.23 / 0.22 & 0.19 / 0.17 & 0.09\\
\hline
Flamenco & 0.20 / 0.18 & 0.16 / 0.19 & 0.15 / 0.16 & 0.16 / 0.17 & 0.17 / 0.17 & 0.12\\
\hline
Foxtrot & 0.22 / 0.24 & 0.23 / 0.24 & 0.21 / 0.21 & 0.16 / 0.18 & 0.20 / 0.22 & 0.12\\
\hline
Latin & 0.23 / 0.23 & 0.19 / 0.20 & 0.21 / 0.22 & 0.20 / 0.19 & 0.21 / 0.21 & 0.14\\
\hline
Quickstep & 0.21 / 0.20 & 0.14 / 0.12 & 0.19 / 0.19 & 0.21 / 0.16 & 0.19 / 0.17 & 0.09\\
\hline
Square & 0.28 / 0.26 & 0.34 / 0.29 & 0.30 / 0.26 & 0.30 / 0.29 & 0.30 / 0.27 & 0.16\\
\hline
Swing & 0.22 / 0.21 & 0.22 / 0.22 & 0.22 / 0.23 & 0.24 / 0.26 & 0.23 / 0.23 & 0.15\\
\hline
Tango & 0.28 / 0.29 & 0.39 / 0.37 & 0.34 / 0.38 & 0.31 / 0.33 & 0.33 / 0.34 & 0.17\\
\hline
Waltz & 0.52 / 0.51 & 0.35 / 0.35 & 0.38 / 0.31 & 0.48 / 0.41 & 0.43 / 0.40 & 0.15\\
\hline
Average & 0.28 / 0.27 & 0.25 / 0.24 & 0.27 / 0.25 & 0.27 / 0.25 & 0.26 / 0.25 & 0.13\\
\hline
Overall & 0.28 / 0.27 & 0.27 / 0.26 & 0.27 / 0.26 & 0.28 / 0.26 & 0.28 / 0.26 & 0.14\\
\hline
\end{tabular}
\end{center}
\end{table*}

\section{Discussion}
\label{sec:discussion}

Our proposed model successfully captured semantics for music and dance, as evidenced by the quantitative evaluation results, which are validated by statistical significance testing, for both instance- and class-level scenarios. Instance-level evaluation confirms that our proposed model is able to generalize the cross-modal features which connect both modalities. This means the model effectively learned how people can move according to the sound of music, as well as how music can sound according to the movement of human bodies. Class-level evaluation further strengthens this conclusion by showing the same effect from a style-based perspective, i.e., the model learned how people can move according to the music style of a song, as well as how music can sound according to the dance style of the movement of human bodies. This result is particularly interesting because the design of both the model and experiments does not explicitly address style, that is, there is no style-based supervision. Since semantic labels are inferenced by humans based on semiotic aspects, this implies that some of the latent semiotic aspects learned by our model are also relevant for these semantic labels, i.e., these aspects are semantically rich. Therefore, modeling semiotic correlations, between audio and dance, effectively uncovers semantic aspects.

The results show a link between musical meaning and kinetic meaning, providing further evidence for embodied cognition semantics in music. This is because embodied semantics ultimately defends that meaning in music is grounded in motor and somatosensory concepts, i.e., movement, physical forces, and physical tension. By observing that dance, a body expression proxy for how those concepts correlate to the musical experience, is semiotically correlated to music artifacts, we show that music semantics is kinetically and biologically grounded. Furthermore, our quantitative results also demonstrate an effective technique for cross-modal retrieval between music audio and dance video, providing the basis for an automatic music video creation tool. This basis consists of a model that can recommend the song that best fits a particular dance video and the dance video that best fits a particular song. The class-level evaluation also validates the whole ranking of results, which means that the model can actually recommend several songs or videos that best fit the dual modality.

\section{Conclusions and future work}
\label{sec:conclusions}

We proposed a computational approach to model music embodied semantics via dance proxies, capable of recommending music audio for dance video and vice-versa. Quantitative evaluation shows this model to be effective for this cross-modal retrieval task and further validates claims about music semantics being defined by embodied cognition. Future work includes correlating audio with 3D motion capture data instead of dance videos in order to verify whether important spatial information is lost in 2D representations, incorporating Laban movement analysis features and other audio features in order to have fine-grained control over which aspects of both music and movement are examined, test the learned semantic spaces in transfer learning settings, and explore the use of generative models (such as \acp{GAN}) to generate and visualize human skeleton dance videos for a given audio input.

\bibliographystyle{model5-names}
\bibliography{musement}

\end{document}